\pdfoutput=1
\documentclass[11pt]{article}

\usepackage[final]{acl}

\usepackage{times}
\usepackage{latexsym}
\usepackage{placeins}
\usepackage{amssymb} % for checkmarks
\usepackage{listings}

\usepackage[T1]{fontenc}
\usepackage[utf8]{inputenc}

\usepackage{microtype}

\usepackage{inconsolata}

\usepackage{graphicx}
\usepackage{afterpage}
\usepackage{booktabs}
\usepackage{hyperref}
\usepackage{url}
\usepackage{multirow}
\usepackage{subfigure}
\usepackage{makecell} 
\usepackage{booktabs}
\usepackage{booktabs} % For formal tables
\usepackage{multirow} % For multirow feature
\usepackage{amsmath} % For \text command
\usepackage{graphicx} % Required for including images

\usepackage{placeins}
\usepackage{xcolor}
\usepackage{booktabs}
\usepackage{multirow}
\usepackage{tabularx}

\definecolor{cvprblue}{rgb}{0.21,0.49,0.74}
\definecolor{citecolor}{HTML}{0071BC}
\definecolor{linkcolor}{HTML}{ED1C24}
\usepackage[capitalize]{cleveref}
\crefname{section}{Sec.}{Secs.}
\crefname{table}{Table}{Tables}
\crefname{figure}{Fig.}{Figs.}
\usepackage[most]{tcolorbox}
\usepackage{float}
\usepackage{xspace}
\tcbset{
  aibox/.style={
    width=474.18663pt,
    top=10pt,
    colback=white,
    colframe=black,
    colbacktitle=black,
    enhanced,
    center,
    attach boxed title to top left={yshift=-0.1in,xshift=0.15in},
    boxed title style={boxrule=0pt,colframe=white,},
  }
}
\newtcolorbox{AIbox}[2][]{aibox,title=#2,#1}

\usepackage{colortbl}

\usepackage{multicol} % Add this for manual column control
\usepackage[most]{tcolorbox}
\usepackage{float}
\usepackage{xspace}
\usepackage{fontsize}
\usepackage{setspace}
\usepackage{geometry}

\usepackage{listings}
\definecolor{codegreen}{rgb}{0,0.6,0}
\definecolor{codegray}{rgb}{0.5,0.5,0.5}
\definecolor{codepurple}{rgb}{0.58,0,0.82}
\definecolor{backcolour}{rgb}{0.95,0.95,0.92}
\definecolor{mypurple}{RGB}{200,192,248}
\definecolor{mypurpledeep}{RGB}{142,126,240}
\definecolor{mygreen}{RGB}{117,170,156}
\definecolor{myyellow}{RGB}{255,192,0}
\definecolor{myblue}{RGB}{57,143,255}
\definecolor{mygrey}{RGB}{231,230,230}
\definecolor{codey}{RGB}{220,220,170}
\definecolor{coder}{RGB}{206,145,120}
\definecolor{codeb}{RGB}{156,220,254}
\definecolor{codenum}{RGB}{204,204,204}

\lstdefinestyle{mystyle}{
    backgroundcolor=\color{backcolour},   
    commentstyle=\color{codegreen},
    keywordstyle=\color{magenta},
    numberstyle=\tiny\color{codegray},
    stringstyle=\color{codepurple},
    basicstyle=\footnotesize,
    breakatwhitespace=false,         
    breaklines=true,                 
    captionpos=b,                    
    keepspaces=true,                 
    numbers=left,                    
    numbersep=5pt,                  
    showspaces=false,                
    showstringspaces=false,
    showtabs=false,                  
    tabsize=2
}

\usepackage{algorithm} % Include the algorithm package
\usepackage{todonotes}
\usepackage{float}
\author{Sophie Ostmeier\\
  \texttt{sostm@stanford.edu} \\\And
  Justin Xu\\
  \texttt{xujustin@stanford.edu} \\\And
  Zhihong Chen\\
  \texttt{zhihongc@stanford.edu} \\
  \AND
  Maya Varma\\
  \texttt{mvarma2@stanford.edu} \\\And
  Louis Blankemeier\\
  \texttt{lblankem@stanford.edu} \\
    \AND
    Christian Bluethgen\\
  \texttt{bluethgen@stanford.edu} \\\And
    Arne Edward Michalson\\
  \texttt{arne64@stanford.edu} \\\And
    Michael Moseley\\
  \texttt{moseley@stanford.edu} \\
    \AND
    Curtis Langlotz\\
  \texttt{langlotz@stanford.edu} \\\And
    Akshay S Chaudhari \thanks{co-senior authorship}\\
  \texttt{akshaysc@stanford.edu} \\\And  
  Jean-Benoit Delbrouck \footnotemark[1]\\
  \texttt{jbdel@stanford.edu} \\
  }
\begin{document}
\title{GREEN: Generative Radiology Report Evaluation and Error Notation}

\maketitle

\begin{abstract}
Evaluating radiology reports is a challenging problem as factual correctness is extremely important due to the need for accurate medical communication about medical images. Existing automatic evaluation metrics either suffer from failing to consider factual correctness (e.g., BLEU and ROUGE) or are limited in their interpretability (e.g., F1CheXpert and F1RadGraph). In this paper, we introduce GREEN (Generative Radiology Report Evaluation and Error Notation), a radiology report generation metric that leverages the natural language understanding of language models to identify and explain clinically significant errors in candidate reports, both quantitatively and qualitatively. Compared to current metrics, GREEN offers: 1) a score aligned with expert preferences, 2) human interpretable explanations of clinically significant errors, enabling feedback loops with end-users, and 3) a lightweight open-source method that reaches the performance of commercial counterparts. We validate our GREEN metric by comparing it to GPT-4, as well as to error counts of 6 experts and preferences of 2 experts. Our method demonstrates not only higher correlation with expert error counts, but simultaneously higher alignment with expert preferences when compared to previous approaches. We publish the code as a pypi package and datsets \footnote{https://stanford-aimi.github.io/green.html}. 
\end{abstract}

\section{Introduction}
\begin{figure*}
    \centering
    \includegraphics[width=0.95\textwidth]{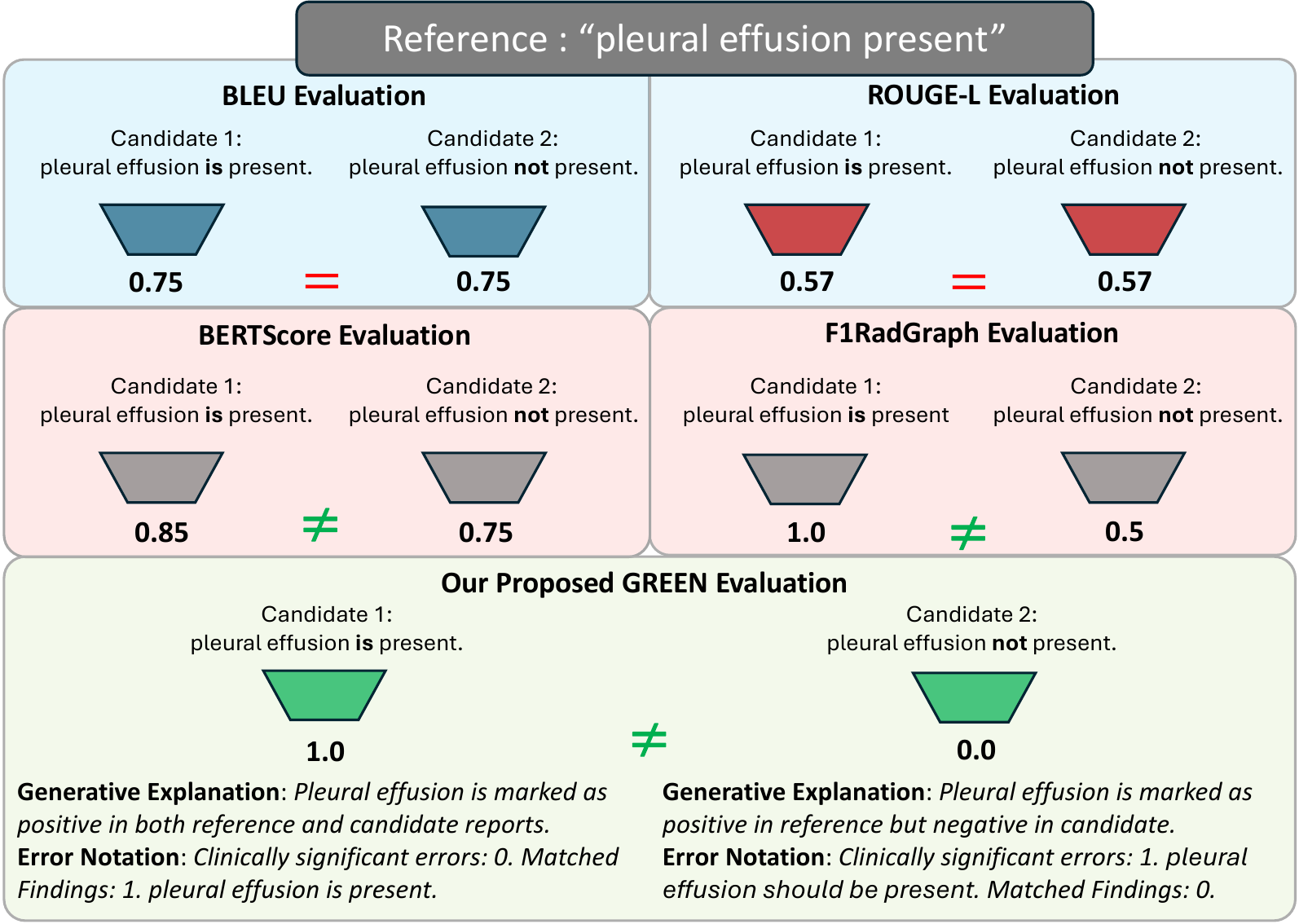}
    \caption{Motivation of GREEN.}
    \label{fig:advantages}
\end{figure*}

Machine learning has enabled great progress in the automatic interpretation of images, where vision language models (VLMs) translate features of images into text~\cite{radford2021learning, liu2024visual}. % clip and llava 
In the medical domain, patient images are interpreted by radiologists, which is referred to as radiology report generation (RRG).
Automated and high-quality RRG has the potential to greatly reduce the workload of radiologists alleviating burdens arising from shortage of radiologists, generally improve clinical communication~\cite{kahn2009toward}, and increase the accuracy of radiology reports~\cite{rajpurkar2023current}. 

Commonly used evaluation metrics in RRG literature~\cite{lin2004rouge, zhang2019bertscore,smit2020combining, delbrouck-etal-2022-improving} seek to evaluate a generated radiology report against a reference report written by a radiologist by leveraging simple n-grams overlap, general language similarity, pathology identification within specific imaging modalities and disease classes, and commercially-available large language models.
To achieve performance on par with radiologists, evaluation metrics must be adept with the radiology language in order to accurately assess factual correctness and levels of uncertainties. Additionally, RRG metrics should be interpretable in a scalable fashion to enable a feedback loop between the generated reports and the experts who review them. Moreover, these metrics should be open-source to allow for assessment of private datasets that require the safeguarding of patient information.

Current RRG metrics fall short of capturing the nuanced and multifaceted nature of radiology reports. To mitigate the current gaps in appropriate metrics for RRG, we introduce \textbf{GREEN (Generative Radiology Evaluation and Error Notation)}. 
The GREEN metric introduces five major contributions:

\begin{itemize}
    \item \textbf{Score}: We introduce and validate a score, which ranges from 0 for the weakest assessment, to 1, marking the highest score achievable. We show that GREEN is adept with radiology language and can accurately assess factual correctness and levels of uncertainties that surpass prior approaches.
    \item \textbf{Interpretable Evaluation Summary}: We provide a method to generate a clear, human-readable evaluation summary independent of the test set size. 
    By providing detailed error categorization with explanations, GREEN enables machine learning practitioners and experts to pinpoint areas for improvement in their trained systems.
    \item \textbf{Practicability}: We open-source the GREEN model that leverages a < 7B parameter language model with similar report evaluation abilities as larger counterparts. 
    This approach decreases GPU requirements and enhances processing speed.
    \item \textbf{Applicability}: Leveraging high-performing commercially-available large language model (LLM) services typically requires a de-identification procedure and institutional review board approval for protected health information (PHI). GREEN is free, open-source, and designed for use in confidential datasets without patient privacy concerns.
    \item \textbf{Multimodality}: GREEN is designed to understand a wide array of pathologies, linguistic styles, and terminologies. We demonstrate that GREEN exhibits a generalized understanding of medical language that spans various imaging modalities and anatomical structures on out-of-distribution (OOD) data, specifically by examining its application to abdominal computed tomography (CT) reports in a zero-shot fashion.
    \item \textbf{Datasets}:
    Lastly, we share the dataset used to develop our models. This dataset encompasses 100,000 annotations from GPT-4 related to chest X-rays (spanning various datasets) and 50,000 annotations across a diverse set of imaging modalities. By making these resources available, we hope to facilitate further research and improvement in the accuracy and reliability of automated radiology report generation systems.
\end{itemize}

\section{Related Work}

The literature demonstrates various advances in generating radiology reports from medical images~\cite{ramesh2022improving, jeong2024multimodal, li2023dynamic, yang2022knowledge, nguyen2021automated, chen2024chexagent, chaves2024training}.
For instance, a set of evaluation metrics are commonly utilized to assess the quality of the generated reports and focus on lexical similarity (e.g., ROUGE-L~\cite{lin2004rouge} and BLEU~\cite{papineni2002bleu}) and factual correctness (e.g., F1CheXbert~\cite{smit2020combining} and F1RadGraph~\cite{delbrouck-etal-2022-improving}). F1CheXbert assesses the accuracy of identified disease labels in reports against a narrow reference, covering only 14 CheXbert classes of common-but-specific chest x-ray findings. F1RadGraph enhances factual correctness evaluations by comparing the agreement on anatomical and observational entities between candidate reports and reference reports, using a graph model trained on human annotations. However, the correlation of F1RadGraph with manual evaluations by radiologists is low, leading to the development of more closely-aligned metrics such as RadCliQ~\cite{yu2023evaluating}. RadCliQ consists of an ensemble of ROUGE, BLEU, CheXbert embedding similarities, and RadGraph to form a composite metric which aims to match expert-generated error counts. While RadCliQ is effective in mirroring these error counts, it has low interpretability as the individual metric weights are unknown and the single numerical score is inadequate for clinical integration (Figure~\ref{fig:advantages}).\\

Our approach stands out from previous metrics by emphasizing clinical relevance and interpretability, showing higher alignment with expert error counts and expert preferences while still leveraging an open-source <7B parameter model.

\section{GREEN} \label{sec:green}
GREEN (Generative Radiology Evaluation and Error Notation) involves \textbf{three primary components}.\\

First, we describe the construction of our \textbf{generative language model}, which is designed to identify and classify errors in radiology reports into six categories (Section~\ref{sec:generative_llm}). This section is subdivided into a process overview, collection of the reference and synthetic candidate reports (Section~\ref{sec:xraydatacollection}) and details of the training process (Section~\ref{sec:green_training}). Second, we elaborate on the \textbf{GREEN score} in Section~\ref{sec:green_score}, including its rationale and significance. Third, we explain the text-form \textbf{GREEN summary} and its usefulness. 

We then outline the steps we took to validate the effectiveness and relevance of GREEN, both quantitatively and qualitatively, in Section~\ref{sec:validation_green}.

\begin{figure}[t]
    \centering
    \includegraphics[width=0.98\linewidth]{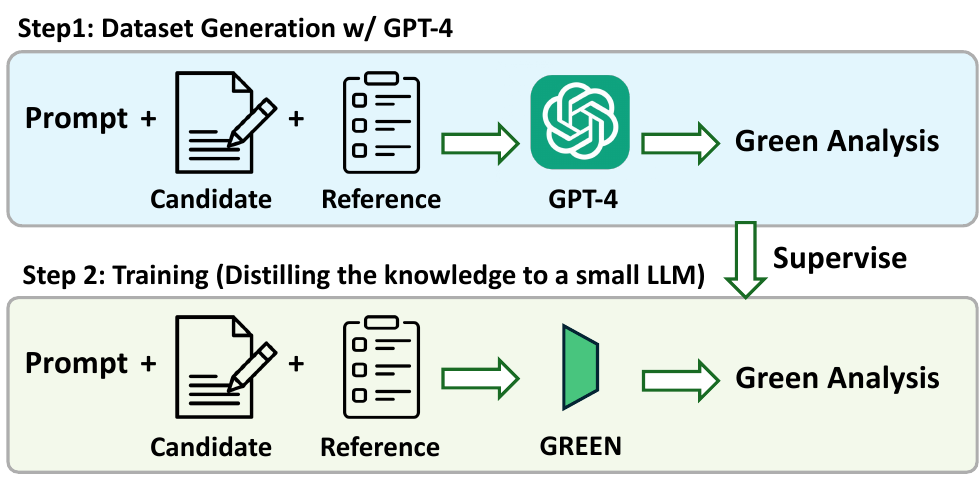}
    \caption{Training procedure of the GREEN model.}
    \label{fig:training_green}
\end{figure}

\subsection{Generative Large Language Model} \label{sec:generative_llm}
We designed a prompt for GPT-4 and LLM-fintuning consisting of the a process overview, criteria for judgement (six distinct error types for each clinically significant and insignificant errors), a reference and a synthetically modified candidate report and a response template (See prompt at \ref{app:gpt4prompttraining} and response at \ref{app:gpt4radgraphpermutation}, Figure \ref{fig:training_green}). Second, we generated a dataset by prompting GPT-4 with the prompt and received a response, called GREEN analysis. Lastely, we used the GREEN analysis to obtain and validate the GREEN score (numerical) and GREEN summary (free text).

\subsubsection{Reference and Synthetic Candidate Reports} \label{sec:xraydatacollection}
To compile pairs of reference and candidate reports, we selected 100,000 reference and generated candidate report pairs from six publicly-available de-identified chest X-ray datasets: MIMIC-CXR~\cite{johnson2019mimic}, MIMIC-PRO~\cite{ramesh2022improving}, CandidPTX~\cite{feng2021curation}, PadChest~\cite{bustos2020padchest}, BIMCV-covid19~\cite{vaya2020bimcv} and OpenI~\cite{demner2012design}. We employed the prompt shown in Appendix~\ref{app:gpt4prompttraining} to task GPT-4 to identify and categorize differences in natural language across six unique clinically-defined categories detailed previously~\citet{yu2023radiology}.

\begin{figure}[!h]
    \centering
    \includegraphics[width=\linewidth]{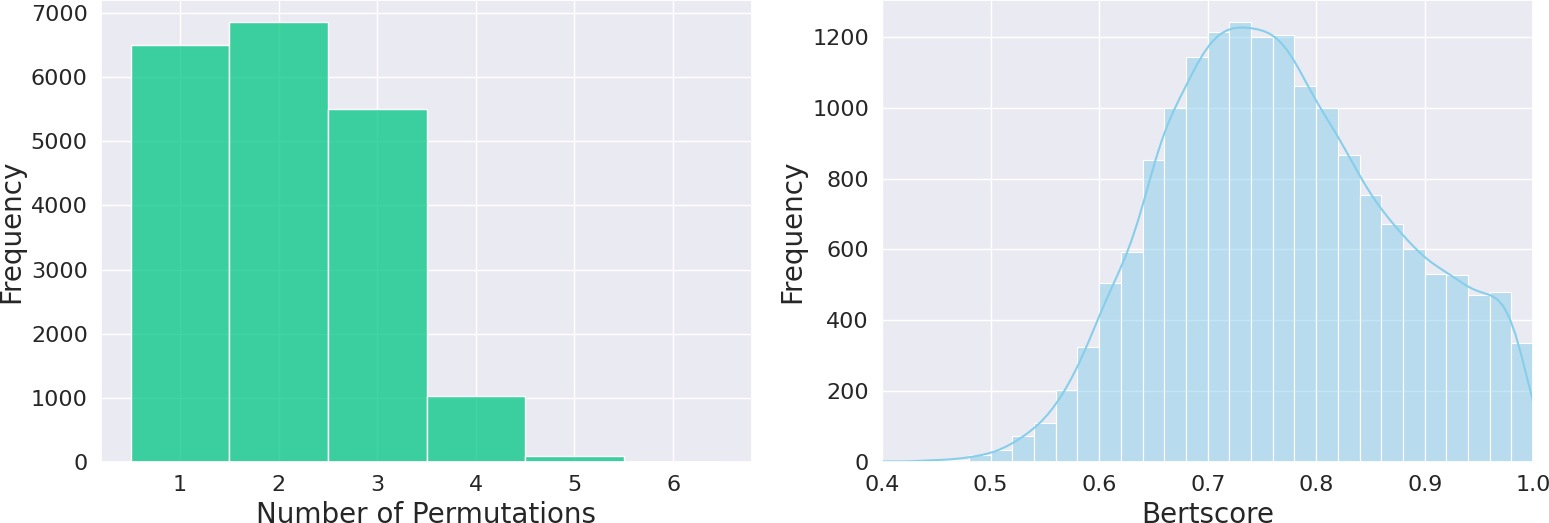}

    \caption{Number of RadGraph permutations among the candidates for 20,000 pairs (left) and BERTScore distribution across 20,000 pairs (right).}
    \label{fig:berts_radg}
\end{figure}

The pairing process used five different heuristics, generating 20,000 unique pairs for each heuristic: i) randomly matching candidates and references, ii) modifying the candidate by removing and shuffling sentences from the original report, iii) using a trained RRG model to create the candidate based on the referenced image, iv) pairing candidates with the closest semantically similar report assessed using BERTScore~\cite{zhang2019bertscore}, and v) creating candidates through RadGraph (named-entity recognition dataset)~\cite{radgraph1.0} permutations of the reference reports, incorporating changes to the presence of findings or by making modifications to size, severity, or location throughout the reports. The number of RadGraph permutations and the BERTScore distribution of the pairs are presented in Figure~\ref{fig:berts_radg}. Additionally, a sample GPT-4 response of a pair with a candidate that includes exactly one RadGraph permutation can be found in Appendix~\ref{app:gpt4radgraphpermutation}.

To maintain uniqueness across the dataset, once pairs were formed using one heuristic, they were excluded from consideration in others, ensuring each of the 100,000 pairs is distinct. Overall, 174,329 unique reports were utilized either as references or candidates in this study.

\begin{figure*}[h]
\begin{AIbox}{GREEN Summary}
\begin{verbatim}
[Summary]:
Green score: mean 0.23 std 0.04

[Clinically Significant Errors]: <accuracy> 
<most representative error>

(a) False report of a finding in the candidate: 0.9 
[Small right pleural effusion]

(b) Missing a finding present in the reference: 0.7 
[Underlying chronic upper lobe scarring.]

(c) Misidentification of a finding's anatomic location/position: 0.4
[The opacity is in the right lower lobe, not the right upper lobe.]

(d) Misassessment of the severity of a finding: 0.8
[Bilateral pleural effusion]

(e) Mentioning a comparison that isn't in the reference: 0.7 
[The candidate report mentions a discussion between doctors, 
which is not present in the reference report]

(f) Omitting a comparison detailing a change from a prior study: 0.5 
[The candidate report does not mention the absence of disease progression]
\end{verbatim}
\end{AIbox} 
\caption{Sample GREEN summary. For each error subcategory, we provide the most representative error explanations, enabling users to pinpoint areas for improvement for their trained systems.}
\label{fig:summary_response}
\end{figure*}

\subsubsection{Baseline Model and Training} \label{sec:green_training}

To enhance performance with medical data, we pre-trained LLaMA-2 and Phi-2 using a comprehensive set of domain-specific datasets to form RadLLaMA-2 and RadPhi-2. These datasets include MIMIC-IV Radiology Reports~\cite{johnson2023mimic}, MIMIC-IV Discharge Summaries, MIMIC-CXR Radiology Reports, and a variety of sources from PubMed (Abstracts and Patient Reports). Specialized datasets such as Wiki Medical Terms\footnote{\url{www.huggingface.co/datasets/gamino/wiki\_medical\_terms}}
and Medical Guidelines\footnote{\url{www.huggingface.co/datasets/epfl-llm/guidelines}}~\cite{vashishth2021improving} were also used.

To obtain a local RRG evaluator for the GREEN metric model, we opted to train open-source models instead of relying on API-based models. 
Specifically, we further fine-tuned RadLLaMA-2 and RadPhi-2, as well as other models of different sizes, architectures, and pre-training datasets, such as LLaMA-2~\cite{touvron2023llama}, Phi-2~\cite{javaheripi_phi2_2023}, and Mistral-v0.1~\cite{jiang2023mistral} (Figure~\ref{fig:training_green}). Models were trained on 8x NVIDIA A100 Tensor Core GPUs with 40GB VRAM using the Huggingface framework with Flash Attention 2, DeepSpeed Stage 3, and the AdamW optimizer. An effective batch size of 2,048 was used for 12 epochs, as well as a base learning rate of 1e-4, a warm-up ratio of 0.05, and a weight decay of 0.1. Training for 7B-parameter models averaged 40 GPU hours, while the 2.7B-parameters models averaged 28 GPU hours. For fast and reliable inference, we employed data parallelism and deterministic sampling with a maximum token length of 2,048 to ensure the reproducibility of the GREEN metric.

\subsection{GREEN Score} \label{sec:green_score}
The GREEN score rewards matched findings, balancing the penalties for clinically significant errors by employing an inverse structure. Consequently, as reports become more concise and accurate, achieving a higher score becomes increasingly challenging.

We employed regular expressions (regex) to parse the counts of errors from the model's output. Specifically, we denoted the count of each type of error as $\text{\# error}_{s,i}$, where the error's clinical significance $s \in \{\mathrm{sig.},\mathrm{insig.}\}$ and subcategory $i \in \{(a),(b),\dots,(f)\}$.

To calculate the GREEN score, we prioritized $\text{\# error}_{\mathrm{sig.},i}$ (errors with the potential to alter clinical decision-making processes) alongside the counts of accurate matched findings, \text{\# matched findings}, for inversion.
The formula for the GREEN score is then expressed as:
\begin{align}
\text{GREEN} = \frac{\text{\# matched findings}}{\text{\# matched findings} + \sum\limits_{i=(a)}^{(f)} \text{\# error}_{\mathrm{sig.},i}}
\end{align}
if \# matched findings > 0, otherwise 0.
Thus, the GREEN score ($\uparrow$) is bounded between 0 and 1.

\subsection{GREEN Summary}
To the best of the authors knowledge, we present the first method for a detailed free text analysis of error explanation per error subcategory for Chest-Xray report generation.
The GREEN analysis (model response) contains three parts: the explanation, clinically significant and insignificant error counts, matched findings. We use clinically significant error counts which would alter the clinical workflow. This part is further divided into six subcategories. The model response for each subcategory includes the error counts and an explanation sentence about each error. We gather these explanation sentences, $e_i \in E_i$, for each subcategory $i$ and embed them using Sentence Transformers~\cite{ni2021large}. 
The embeddings are then grouped into $k$ cluster, where $k$ is determined by the silhouette distance~\cite{shahapure2020cluster}. We choose the largest cluster and the closest to the mean embeddings of the largest cluster. 
We map these embeddings back to the respective individual sentences, which are then included in the GREEN summary (Figure~\ref{fig:summary_response}).
For example, in Figure~\ref{fig:summary_response}, subcategory $(a)$ has "\textit{(Small right pleural effusion, Small right pleural effusion, Small right pleural effusion)}" as the three closest members to the largest cluster, which indicates significant hallucinations of a "\textit{small right pleural effusion}". This opens up the possibility for targeted detection of data biases or quality issues, as well as specific areas for model improvement. 

\subsection{Validation}\label{sec:validation_green}

\subsubsection{Expert Error Counts Dataset} \label{sec:rexval}
To validate the GREEN score in a clinical setting, we utilized the publicly-available ReXVal dataset, which includes assessments from six board-certified radiologists on 200 pairs of generated radiology reports from 50 cases of the MIMIC-CXR test set~\cite{yu2023radiology}. 
Each radiologist counted the occurrences of six specific error types denoted earlier, distinguishing between errors of clinical significance and those considered insignificant. 
%This dataset does not include matched findings and an explanation of the errors. 

\subsubsection{Expert Preference Dataset} \label{sec:preferences}
Since we lack a ground-truth score for evaluating radiology reports, we turned to a radiologist preference dataset. This helped to address the shortcomings of just comparing errors counts in the previous section, may reduce score overfitting to the RexVal dataset.  
In particular, when radiologists compare two reports, they do so with an intuitive weighting of matched findings and significant and insignificant errors. 
As such, the preference dataset enabled us to determine which metric most effectively replicates expert evaluations and allowed us to assess GREEN as a preference generator ~\cite{rafailov2024direct, ethayarajh2024kto}.

We collected 100 pairwise preferences by two board-certified radiologists (with over 5 and 25 years of experience). The dataset comprised of 50 cases of the ReXVal dataset~\cite{yu2023radiology}, supplemented by an additional 50 cases randomly selected from the MIMIC-CXR test set. The two radiologists were presented with a chest X-ray alongside two corresponding candidate reports generated by an image-captioning model fine-tuned on the MIMIC CXR training set\footnote{\url{https://huggingface.co/nlpconnect/vit-gpt2-image-captioning}}. The primary task for the radiologist was to select the candidate report they preferred and to quantify their confidence in this selection on a scale ranging from 1 to 10. The intention behind this was to categorize the complexity of the task. In essence, when radiologists have a high degree of confidence in their chosen report for a given case, it is anticipated that the automated preference generator will show the highest concordance, as the task is deemed less challenging.
The quality of the two candidates is considered to be equal when experts had differing preferences, hence implying no preference. We then excluded such cases from consideration.
\begin{table}[ht]
\centering
\caption{Difference between ReXVal experts and the GREEN model measured using mean absolute error of significant error counts.}
\resizebox{\linewidth}{!}{
\begin{tabular}{cccccccc} 
\toprule
& $\mathbf{0}$ & $\mathbf{1}$ & $\mathbf{2}$ & $\mathbf{3}$ & $\mathbf{4}$ & $\mathbf{5}$ & GREEN \\ \midrule
$\mathbf{0}$ & $-$ & 0.505 & 0.835 & 0.675 & 0.495 & 1.130 & 1.160 \\
$\mathbf{1}$ & 0.505 & $-$ & 1.100 & 0.870 & 0.660 & 1.365 & 1.485 \\
$\mathbf{2}$ & 0.835 & 1.100 & $-$ & 0.730 & 0.770 & 0.725 & 0.715 \\
$\mathbf{3}$ & 0.675 & 0.870 & 0.730 & $-$ & 0.570 & 0.965 & 0.895 \\
$\mathbf{4}$ & 0.495 & 0.660 & 0.770 & 0.570 & $-$ & 1.025 & 1.005 \\
$\mathbf{5}$ & 1.130 & 1.365 & 0.725 & 0.965 & 1.025 & $-$ & 0.930 \\
GREEN & 1.160 & 1.485 & 0.715 & 0.895 & 1.005 & 0.930 & $-$\\
\bottomrule
\label{tab:inter_expert_mae}
\end{tabular}}
\end{table}

\section{Experiments}

\subsection{Inter-Expert Analysis} \label{sec:interexpert_analysis}
The baseline for model performance was established by comparing the correlation between experts to each other from the ReXVal dataset using Kendall's Tau coefficient~\cite{yu2023radiology}. The correlation between the 6 experts was less than the average correlation across experts, which spans from 0.41 to 0.60 (Appendix Figure~\ref{fig:inter-expert_corr}). 
Additionally, we assessed the discrepancy between experts by computing the mean absolute error of significant error counts, resulting in a mean difference of 0.83 ± 0.13 (Table~\ref{tab:inter_expert_mae}). These inter-expert measures serve as upper bound for the GREEN model performance outlined below.
For testing the difference in location of the mean difference between expert, we calculate the statistics using the Wilcoxon-signed rank test. These values show that the experts themselves already significantly differ, leading us to the conclusion that the Wilcoxon-signed rank test is less suitable for comparison of error counts between GREEN, GPT, and experts (Appendix Tables \ref{tab:rater_stats} and \ref{tab:model_stats}).

\subsection{Performance on Training Data Distribution}

We measured performance of the GREEN metric models by sampling deterministically and comparing the mean absolute errors and classical lexical metrics against reference labels from GPT-4. We found that RadPhi-2 and RadLLaMA-2 exhibit the lowest mean absolute difference for clinically significant errors of 0.63 ± 0.99 (Table \ref{tab:internal_MAE}) and the highest classical lexical metrics with a mean BERTScore of 0.84 ± 0.10 (Table~\ref{tab:internal_lexical}). We measured clustering and summary consistency with language similarity metrics like BERTScore. Consistent with the quantitative results, we found that RadPhi-2 and RadLLaMA-2 yielded the best natural language agreement with GPT-4 (Table~\ref{tab:internal_MAE}).
\begin{table*}[]
\centering
\caption{Results on the internal test set (10,000 examples) when compared to GPT-4 error counts. $^1$Mean absolute error ± standard deviation, $^2$Average error count.}
\label{tab:internal_MAE}
\resizebox{\linewidth}{!}{%
\begin{tabular}{cccccc}
\toprule
 \multirow{3}{*}{Language Model} & \multicolumn{3}{c}{MAE ± STD $^1$} &  \multicolumn{1}{c}{$\Delta$ GREEN $\downarrow$} \\ \cline{2-5} 
& \makecell{$\Delta$Sig. Error Count } $\downarrow$ & \makecell{$\Delta$Insig. Error Count } $\downarrow$& $\Delta$Matched Findings $\downarrow$ &\\ 
 & 3.1\scriptsize ± 2.6$^2$  & 0.15 \scriptsize ± 0.52$^2$ & 2.07  \scriptsize ± 1.84$^2$\\ \midrule
Mistral-v0.1 (7B) & 0.97 \scriptsize ± 1.18 & 0.22 \scriptsize ± 0.58 & 0.44 \scriptsize ± 0.70 & 0.11  \scriptsize ± 0.17 \\
 LLaMA-2 (7B)&  1.35 \scriptsize ± 1.40&  \cellcolor[HTML]{C0C0C0}\bfseries 0.15 \scriptsize ± 0.52  & 1.62 \scriptsize ± 1.67 & 0.29 \scriptsize ± 0.26 \\
Phi-2 (2.7B) & 0.84 \scriptsize ± 1.14& 0.20 \scriptsize ± 0.58 & 0.34 \scriptsize ± 0.59 & 0.09 \scriptsize ± 0.14\\
RadLLaMA-2 (7B) & \cellcolor[HTML]{EFEFEF}0.70 \scriptsize ± 0.99 & 0.20 \scriptsize ± 0.57 & \cellcolor[HTML]{EFEFEF}0.29 \scriptsize ± 0.54 & \cellcolor[HTML]{EFEFEF}0.08 \scriptsize ± 0.13\\
 RadPhi-2 (2.7B) & \cellcolor[HTML]{C0C0C0}\bfseries 0.63 \scriptsize ± 0.99 & \cellcolor[HTML]{EFEFEF}0.18  \scriptsize ± 0.57& \cellcolor[HTML]{C0C0C0}\bfseries0.26  \scriptsize ± 0.53 &  \cellcolor[HTML]{C0C0C0}\bfseries0.06 \scriptsize ± 0.12\\
\bottomrule
\end{tabular}
}
\end{table*}

\begin{table}[h]
\centering
\caption{Results on the Internal test set (10,000 examples) when compared to GPT-4 responses.}
\resizebox{\linewidth}{!}{%
\begin{tabular}{ccccc}
\toprule
 \multirow{2}{*}{Language Model}  &\multicolumn{3}{c}{Lexical} \\ \cline{2-4} 
& BERTScore$\uparrow$ &    ROUGE-L $\uparrow$&  BLEU $\uparrow$ \\ \midrule
Mistral-v0.1 (7B)&    0.80 \scriptsize ± 0.11 & 0.68 \scriptsize± 0.18 & 0.54 \scriptsize ± 0.22\\
LLaMA-2 (7B) &    0.78 \scriptsize ± 0.12& 0.65 \scriptsize ± 0.19 & 0.53 \scriptsize ± 0.21 \\
Phi-2 (2.7B) &0.80 \scriptsize ± 0.11 & 0.70 \scriptsize ± 0.18 & 0.54 \scriptsize ± 0.23 \\
RadLLaMA-2 (7B) &  \cellcolor[HTML]{EFEFEF}0.83 \scriptsize ± 0.24 &\cellcolor[HTML]{EFEFEF}0.73 \scriptsize ± 0.17 & \cellcolor[HTML]{EFEFEF}0.59 \scriptsize ± 0.23\\
 RadPhi-2 (2.7B)  &\cellcolor[HTML]{C0C0C0}\bfseries  0.84 \scriptsize ± 0.10 & \cellcolor[HTML]{C0C0C0}\bfseries 0.76 \scriptsize ± 0.17 & \cellcolor[HTML]{C0C0C0}\bfseries 0.64 \scriptsize ± 0.23 &\\\bottomrule
\end{tabular}
}
\label{tab:internal_lexical}
\end{table}

\subsection{Performance on Validation Data Distribution}

To analyze the performance upper bound of GREEN, we inferred GREEN scores from GPT-4 responses on the validation set, and referred to it as GREEN-GPT-4.

\subsubsection{Expert Error Counts}
To quantitatively validate GREEN, we measured the mean absolute difference and accuracy relative to the average radiologist, as detailed in Section~\ref{sec:interexpert_analysis}. We found that, overall, RadLLaMA-2 exhibits the lowest differences to the mean radiologist's error counts (1.54 ± 1.36 sig. error difference), which approaches the performance of GPT-4 (1.51 ± 1.29 sig. error difference).
Compared to all experts individually, RadLLaMA-2 exhibits an average difference of 1.02 ± 0.27, which is within the boundaries of the average inter-expert difference of 0.83 ± 0.13.
Drawing from these quantitative results, along with the findings presented in Section~\ref{sec:validation_green}, we selected RadLLaMA-2 as the GREEN model for all future experiments (Table
\begin{table*}[h]
\centering
\caption{Results on the external validation set (200 examples) compared to ReXVal human experts. $^1$MAE: Mean Absolute Error of the sum of Sig. Errors and Insig. Errors, $^2$STD: Standard Deviation, $^3$Average error count. (a)-(f) Accuracy for each significant error category: (a) False report of a finding in the candidate, (b) Missing a finding present in the reference, (c) Misidentification of a finding's anatomic location/position, (d) Misassessment of the severity of a finding, (e) Mentioning a comparison that isn't in the reference, and (f) Omitting a comparison detailing a change from a prior study.$^4$PreRad=pretrained on radiology text, Section \ref{sec:green_training}, $^5$MM=multimodal, described in Section \ref{sec:multimodal_data}.}
 \label{tab:fine_grained}
\resizebox{\linewidth}{!}{%
\begin{tabular}{llllcccccccc} % cccccccccc
\toprule
\multirow{2}{*}{Language Model}  & \multicolumn{3}{c}{\multirow{2}{*}{Data}} & \multicolumn{2}{c}{MAE$^1$ ± STD $^2$}& \multicolumn{6}{c}{Accuracy $\uparrow$} \\  \cline{5-12} 
& & &  & \makecell{$\Delta$Sig. Error$\downarrow$ } & \makecell{$\Delta$Insig. Error$\downarrow$ } & (a) & (b) & (c) & (d) & (e) & (f) \\ 
 & PreRad$^4$  & CXR & MM$^5$ & 7.03 \scriptsize ± 1.16 $^3$& 0.47 \scriptsize ± 0.52$^3$\\ \midrule
Mistral-v0.1 (7B) & & \checkmark &  & 2.60 \scriptsize ±  1.91 & 0.87 \scriptsize ±  0.94& 0.13	&0.31&	\cellcolor[HTML]{EFEFEF}0.62&\cellcolor[HTML]{C0C0C0}\bfseries	0.59	&0.48	&0.67 \\
LLaMA-2 (7B) & & \checkmark& & 2.62 \scriptsize ± 1.25 &  \cellcolor[HTML]{C0C0C0}\bfseries 0.47 \scriptsize ± 0.52 & 0.10&	0.23	& \cellcolor[HTML]{C0C0C0}\bfseries0.65	& \cellcolor[HTML]{C0C0C0}\bfseries0.59	& \cellcolor[HTML]{C0C0C0}\bfseries0.68&	\cellcolor[HTML]{C0C0C0}\bfseries0.70\\ 
LLaMA-2 (7B) &  \checkmark&\checkmark & & \cellcolor[HTML]{C0C0C0}\bfseries 1.54 \scriptsize ± 1.36 &   \cellcolor[HTML]{EFEFEF} 0.51 \scriptsize ± 0.54 & \cellcolor[HTML]{C0C0C0}\bfseries 0.34	& \cellcolor[HTML]{C0C0C0}\bfseries 0.38	&0.60	&0.54	&0.65	& \cellcolor[HTML]{EFEFEF} 0.68 \\
Phi-2 (2.7B) & & \checkmark& &  2.10 \scriptsize ± 1.39 & 0.65 \scriptsize ± 0.70 & 
\cellcolor[HTML]{C0C0C0}\bfseries 0.34 & 	0.08	& \cellcolor[HTML]{C0C0C0}\bfseries 0.65	& \cellcolor[HTML]{EFEFEF} 0.57	&\cellcolor[HTML]{EFEFEF}  0.66	& 0.53 \\ 
Phi-2 (2.7B) &\checkmark&\checkmark & &   2.08 \scriptsize ± 1.15 &  0.55  \scriptsize ± 0.61 & 0.19& 	0.18	& \cellcolor[HTML]{EFEFEF}0.62	& \cellcolor[HTML]{EFEFEF} 0.57 & 	0.62& 	0.61 \\
% Phi-3 & &\checkmark & &  \cellcolor[HTML]{EFEFEF} 2.03 \scriptsize ± 1.59 & 0.54 \scriptsize ± 0.56 & 0.31& 	0.3	& \cellcolor[HTML]{EFEFEF}0.62	& \cellcolor[HTML]{C0C0C0}\bfseries 0.59	&0.58& 0.66 \\ 
% Phi-3 & &\checkmark & \checkmark &   2.28 \scriptsize ± 1.22 & 0.66 \scriptsize ± 0.71  & 0.23& 	0.19& 0.6	& 0.57	& 0.64 & 0.26 \\ 
% Gemma-2b & &\checkmark & &  2.29 \scriptsize ± 1.90 & 0.73 \scriptsize ± 0.95 & 0.27	&\cellcolor[HTML]{EFEFEF} 0.30	& 0.61& \cellcolor[HTML]{C0C0C0}\bfseries	0.59	& 0.57& 	0.60 \\ 
% Gemma-2b & & \checkmark&\checkmark &  2.25 \scriptsize ± 1.44& 0.55 \scriptsize ± 0.58 & 0.32 & 	0.28 &	0.57 & 	0.55	& \cellcolor[HTML]{EFEFEF} 0.66	& 0.19\\  \cline{5-12}
GREEN GPT-4 & & & & 1.51 \scriptsize ± 1.29 & 0.52 \scriptsize ± 0.55 & 0.32 &  0.40 &  0.65 &  0.59 &  0.68 &  0.70\\  \midrule
\end{tabular}
 }
\end{table*}).

To validate GREEN against existing metrics, we assessed the correlation between the radiologists total error count (sum of significant and insignificant errors) and the classical metrics, alongside GPT-4 and GREEN as proposed in \cite{yu2023radiology}. Both GREEN-GPT-4 and GREEN demonstrated similar and stronger correlations compared to classical metrics.
We noted that the GREEN correlation significantly outperforms that of RadGraph, despite RadGraph being trained on human annotations (RadGraph: 0.47 (95\% CI, -0.55 0.39) vs. GREEN: 0.63 (95\% CI, 0.69 0.56) )(Table~\ref{tab:correlations}). 

Compared to the inter-expert correlation, GREEN exhibits a competitive degree of correlation at 0.63 compared to the range from 0.48 to 0.64 on the same examples (Figure~\ref{fig:inter-expert_corr}). While this approach enables cross-metric and cross-study comparisons, it may not be clinically optimal. The total error count fails to differentiate between error significance, compares scores on different scales (0 to 1 versus infinity to 0) and newly proposed scores may overfit to this benchmark design.

Furthermore, our analysis revealed that the correlation coefficient between the unweighted total error count of GREEN's (summing clinically significant and insignificant errors) and radiologists' total error counts is 0.79 (95\% CI: 0.74-0.83). This correlation is statistically indistinguishable from the experts total error and the performance of both GPT-4-based G-Rad ~\cite{chaves2024training} and GPT-4-based GREEN when compared to radiologists' total error counts.

For more fine-grained analyses we calculate the error count differences for each subcategory (Table \ref{tab:error_subcategory}). The GREEN model only has one shaded cell for category C, indicating that the largest deviations are “Misidentification of a finding’s anatomic location/position” compared to all 6 experts and GPT-4.
\begin{table}[H]
\centering
\caption{Fine grained error difference to the mean rater, subcategories (a)-(f): Accuracy for each significant error category: (a) False report of a finding in the candidate, (b) Missing a finding present in the reference, (c) Misidentification of a finding's anatomic location/position, (d) Misassessment of the severity of a finding, (e) Mentioning a comparison that isn't in the reference, and (f) Omitting a comparison detailing a change from a prior study.}
\label{tab:error_subcategory}
\resizebox{\linewidth}{!}{%
\begin{tabular}{lcccccc}
\toprule
Rater & (a) & (b) & (c) & (d) & (e) & (f) \\
\midrule
Expert 0 & $0.28 \pm 0.38$ & $0.26 \pm 0.37$ & $0.15 \pm 0.26$ & $0.12 \pm 0.21$ & $0.11 \pm 0.21$ & $0.09 \pm 0.17$ \\
Expert 1 & $0.30 \pm 0.37$ & $0.25 \pm 0.38$ & $0.15 \pm 0.26$ & $0.13 \pm 0.23$ & $0.10 \pm 0.18$ & $0.09 \pm 0.17$ \\
Expert 2 & $0.30 \pm 0.36$ & $0.26 \pm 0.46$ & $0.16 \pm 0.29$ &\cellcolor[HTML]{FFCCCC}  $0.24 \pm 0.40$ & $0.10 \pm 0.19$ & $0.10 \pm 0.20$ \\
Expert 3 & $0.28 \pm 0.36$ & $0.28 \pm 0.39$ & $0.14 \pm 0.25$ & $0.15 \pm 0.27$ & \cellcolor[HTML]{FFCCCC} $0.18 \pm 0.37$ & \cellcolor[HTML]{FFCCCC} $0.18 \pm 0.36$ \\
Expert 4 & $0.34 \pm 0.42$ & $0.22 \pm 0.32$ & $0.16 \pm 0.29$ & $0.16 \pm 0.26$ & $0.15 \pm 0.29$ & $0.14 \pm 0.27$ \\
Expert 5 & \cellcolor[HTML]{FFCCCC} $0.67 \pm 0.77$ & $0.37 \pm 0.55$ & $0.13 \pm 0.23$ & $0.19 \pm 0.34$ & $0.12 \pm 0.22$ & $0.08 \pm 0.15$ \\ \midrule
GREEN      & 0.36 ± 0.45    & 0.38 ± 0.48    & \cellcolor[HTML]{FFCCCC} 0.21 ± 0.34    & 0.23 ± 0.36    & 0.14 ± 0.25    & 0.11 ± 0.21    \\
GREEN GPT-4          & 0.32 ± 0.37    & \cellcolor[HTML]{FFCCCC} 0.43 ± 0.54    & 0.15 ± 0.28    & 0.2 ± 0.3      & 0.13 ± 0.24    & 0.14 ± 0.28    \\
\bottomrule
\end{tabular}
}
\end{table}

\subsubsection{Expert Preferences}
We further used expert preferences to determine whether the summed error counts or the GREEN score best mimics expert evaluation. 
This approach is based on the assumption that clinically significant errors, insignificant errors, and matched findings carry different weights in determining the quality of a candidate report.
\begin{table}
\centering
\caption{Correlation analysis between metrics to total error count of 6 radiologists in the ReXVal dataset (200 examples).}
\begin{tabular}{ll}
\toprule
Metrics &                Kendall's Tau $\uparrow$ \\
\midrule
QAFactEval          &  0.32 \scriptsize(95\% CI, 0.21 0.41) \\
AlignScore          &  0.32 \scriptsize(95\% CI, 0.23 0.41) \\
BLEU                &  0.35  \scriptsize(95\% CI, 0.25 0.43) \\
BERTScore           &  0.49  \scriptsize(95\% CI, 0.40 0.56) \\
F1RadGraph          &  0.57  \scriptsize(95\% CI, 0.49 0.63) \\
ROUGE-L             &  0.56  \scriptsize(95\% CI, 0.48 0.63) \\
RadCliQ-v1          &   0.61 \scriptsize(95\% CI, 0.55 0.66) \\
 GREEN (ours) &   \cellcolor[HTML]{EFEFEF}0.63  \scriptsize(95\% CI, 0.56 0.69) \\
 GREEN GPT-4 (ours) & \cellcolor[HTML]{C0C0C0} \bfseries0.64  \scriptsize(95\% CI, 0.57 0.70) \\ \midrule
Error count G-Rad (GPT-4)& 0.76 \scriptsize(95\% CI, 0.70, 0.80)\\
Error count GREEN & \cellcolor[HTML]{EFEFEF} 0.79 \scriptsize(95\% CI, 0.74 0.83)\\
Error count GPT-4 &\cellcolor[HTML]{C0C0C0} \bfseries 0.79 \scriptsize(95\% CI, 0.75 0.83)\\ \midrule
Inter-Expert & 0.81 \scriptsize(95\% CI, 0.78 0.83)\\
\bottomrule
\end{tabular}
\label{tab:correlations}
\end{table}

The accuracy of the generated preferences was measured by how often the expert-preferred report matched the report that had a higher score from GREEN, a lower summed error count, or the preference of GPT-4. The prompt that was used for GPT-4 preferences is shown in Appendix~\ref{app:gpt4promptpreference}). 

We observed the highest accuracy for GREEN and GREEN-GPT-4, which both outperforms the summed error count approach. The preferences of GPT-4 exhibited an accuracy of 0.23 (95\% CI, 0.13 0.36) (Table~\ref{tab:preferences}).

\begin{figure}[]
    \centering
    \includegraphics[width=0.5\textwidth]{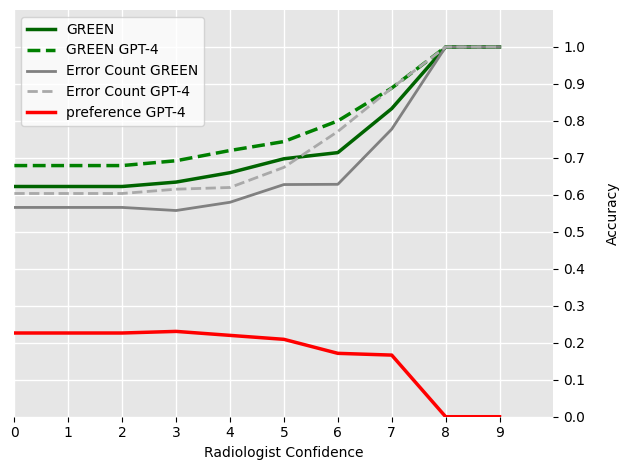}
    \caption{Radiologist confidence vs. accuracy of preference labeling. As the confidence of the experts in their preferences increases, the GREEN score demonstrates the highest alignment with expert preferences as compared to the approach of using just the summed error counts. This difference was quantified using accuracy (green lines). Of note, if GPT-4 is asked directly about a preference, it aligns poorly with the expert preference. However, when the GREEN score formula is applied, a higher accuracy is shown even at lower expert confidence levels. Detailed results can be found in Table~\ref{tab:preferences}.}
    \label{fig:expert_dataset}
\end{figure}

Upon examining the impact of varying confidence levels (Figure~\ref{fig:expert_dataset}), we observed that GREEN's preference alignment improves in conjunction with increased radiologist confidence, distinguishing it from the approach of using just the total error counts or the direct GPT-4 preference with low accuracy.

\begin{table}[h]
\centering
\caption{Accuracies with 95\% CI of various preferences when compared to expert preferences.}
\begin{tabular}{ll} 
\toprule
& Accuracy \\ \midrule
Preference GPT-4 & 0.23 \scriptsize (95\% CI, 0.13 0.36) \\ \midrule
Error Count GREEN & 0.57 \scriptsize (95\% CI, 0.43 0.70) \\
Error Count GPT-4 & 0.60 \scriptsize (95\% CI, 0.47 0.74) \\ \midrule
GREEN (ours)& \cellcolor[HTML]{EFEFEF} 0.62 \scriptsize (95\% CI, 0.49 0.75) \\ 
GREEN GPT-4 (ours)& \cellcolor[HTML]{C0C0C0} \bfseries 0.68 \scriptsize (95\% CI, 0.55 0.79) \\ 
\bottomrule
\label{tab:preferences}
\end{tabular}
\end{table}
\begin{table*}[]
\caption{Adapting GREEN to any imaging modality: Performances on Out-of-Chest X-ray and OOD Data in Zero-Shot and Trained Conditions. $^1$Mean absolute error ± standard variation, $^2$Average Error, $^3$Modalities include X-ray, computed tomography, magnetic resonance imaging, and ultrasound.}
\begin{tabular}{lllllll}
\toprule
\multirow{4}{*}{\makecell{Evaluation\\ data}} & \multirow{4}{*}{\makecell{Training\\ data}} & \multicolumn{3}{c}{MAE ± STD $^1$} & \multicolumn{2}{c}{$\Delta$ GREEN $\downarrow$} \\ 
\cmidrule(lr){3-5} \cmidrule(lr){6-7}
& & \makecell{$\Delta$Sig. Error\\ Count $\downarrow$} & \makecell{$\Delta$Insig. Error\\ Count $\downarrow$} & \makecell{$\Delta$Matched\\ Findings $\downarrow$} & \multicolumn{2}{c}{} \\
\cmidrule(lr){3-5}
& & 2.24 \scriptsize± 2.28 $^2$ & 0.16 \scriptsize± 0.45 $^2$ & 1.83 \scriptsize± 2.25 $^2$ & \multicolumn{2}{c}{} \\
\midrule
\multirow{2}{*}{MIMIC-IV-Notes$^3$} & CXR dataset & 1.05 \scriptsize± 1.51 & 0.20 \scriptsize± 0.51 & 0.49 \scriptsize± 1.00 & 0.10 \scriptsize± 0.17 & \\
& CXR + Out-of CXR dataset & 0.61 \scriptsize± 0.99 & 0.19 \scriptsize± 0.48 & 0.34 \scriptsize± 1.04 & 0.07 \scriptsize± 0.15 & \\
\midrule
 &  & 9.19 \scriptsize ± 3.81 $^2$ & 0.06 \scriptsize ± 0.24 $^2$& 8.56 \scriptsize ± 2.74 $^2$\\ \midrule
\multirow{2}{*}{\makecell{Abdominal CT,\\ OOD}} & CXR dataset & 5.31 \scriptsize± 2.82 & 0.06 \scriptsize± 0.24 & 4.09 \scriptsize± 2.74 & 0.21 \scriptsize± 0.17 & \\
& CXR + Out-of CXR dataset & 3.12 \scriptsize± 2.03 & 0.19 \scriptsize± 0.53 & 3.56 \scriptsize± 3.28 & 0.17 \scriptsize± 0.23 & \\
\bottomrule
\end{tabular}
\end{table*}
\begin{table*}[]
\centering
\caption{Adapting GREEN to any imaging modality: Performance on Out-of-Chest X-ray and OOD data distribution in Zero-Shot and Trained Conditions based on Lexical Metrics. $^1$Modalities include X-ray, computed tomography, magnetic resonance imaging, and ultrasound.}
\label{tab:multimodel_classical}
\begin{tabular}{lllll}
\toprule
\multirow{2}{*}{\makecell{Evaluation\\ data}} & \multirow{2}{*}{\makecell{Training\\ data}} & \multicolumn{3}{c}{Lexical} \\ 
\cmidrule(lr){3-5}
& & BERTScore $\uparrow$ & ROUGE-L $\uparrow$ & BLEU $\uparrow$ \\
\midrule
\multirow{2}{*}{MIMIC-IV-Notes$^1$} & CXR dataset & 0.74 \scriptsize± 0.12 & 0.62 \scriptsize± 0.18 & 0.45 \scriptsize± 0.21 \\
& CXR + Out-of CXR dataset & 0.81 \scriptsize± 0.09 & 0.73 \scriptsize± 0.15 & 0.60 \scriptsize± 0.18 \\
\midrule
\multirow{2}{*}{\makecell{Abdominal CT,\\ OOD}} & CXR dataset & 0.68 \scriptsize± 0.12 & 0.58 \scriptsize± 0.15 & 0.41 \scriptsize± 0.13 \\
& CXR + Out-of CXR dataset & 0.71 \scriptsize± 0.06 & 0.58 \scriptsize± 0.07 & 0.45 \scriptsize± 0.06 \\
\bottomrule
\end{tabular}
\end{table*}

\newpage
\section{Multimodality Generalizability}
We now demonstrate how this method can be applied to various other imaging modalities.

\subsection{Out-of-Chest X-ray Dataset}
\label{sec:multimodal_data}
Recent works extended RRG capabilities of VLMs to other imaging modalities~\cite{hamamci2024foundation, bai2024m3d}.
To extend the GREEN model to new imaging modalities beyond chest X-rays, we created a dataset analogous to the training dataset used for the GREEN chest X-ray (Section~\ref{sec:generative_llm}), but without access to RRG models to generate candidate reports for every modality. We did this to validate our method on a range of imaging modalities for which RRG models may not yet exist.

This new dataset is also based on MIMIC-IV Radiology Reports, which includes 2,321,355 de-identified radiology reports from 237,427 patients. It covers a variety of imaging modalities such as X-ray, computed tomography, magnetic resonance imaging, and ultrasound, as referenced in~\cite{johnson2023mimic}.

We first uniformly sampled reports to maintain a distribution of cases similar to that described in~\cite{johnson2023mimic}. Secondly, we used 4 methods to modify the radiology reports to generate 50,000 candidate reports: i) re-arranging the order of sentences, ii) removing sentences, iii) randomly pairing sentences, and iv) modifying the report by sampling random combinations of error categories and asking GPT-4 to incorporate errors into the reports to generate a candidate report (if no error categories are sampled, GPT-4 is asked to rephrase the report with the same meaning by changing a small number of words) (Appendix \ref{app:changereportprompt}). 

We then prompted GPT-4 to evaluate the differences with the same prompt design as with the chest X-ray data. 
We further split these 50,000 reports into training, validation, and test sets according to the same 80/10/10 ratio and combined them with the initial chest X-ray dataset.

\subsection{OOD External Abdominal CT dataset}
We randomly chose 15 pairs of reference and candidate reports from an abdominal CT dataset. The dataset originated from the Stanford University Medical Center's radiology  dataset, which includes examinations from December 2012 to October 2018.

\subsection{Out-of-Chest X-ray Experiments}
We first evaluated zero-shot performance of the GREEN model on the Out-of-Chest X-ray dataset (1.05 ± 1.51 sig. error count difference) and on the external OOD (5.31 ± 2.82 sig. error count difference).
We fine-tuned the best checkpoint of the GREEN model on the Out-of-Chest X-ray dataset with a batch size of 80 for 8 epochs and the same hyperparameters as mentioned in Section~\ref{sec:green_training}.
We used the same evaluation experiments as in the previous section.
We found that further fine-tuning on multimodality data improves the sig. error count difference and the text similarity metrics for both the in-distribution and out-of-distribution data (0.61 ± 0.99 and 3.12 ± 2.03 sig. error count difference). 

\section{Conclusion}

In this study, we introduced GREEN (Generative Radiology Report Evaluation and Error Notation), a novel metric aimed at enhancing the evaluation of radiology reports. GREEN outperforms existing metrics by aligning closely with the nuanced requirements of medical diagnostics through its precise assessment of factual correctness and uncertainties. The score's high correlation with expert evaluations underscores its effectiveness.

The open-source nature of GREEN supports widespread use and collaborative improvements without compromising data privacy. Its lightweight design ensures practicality across diverse settings, reducing computational demands. Additionally, GREEN’s adaptability across different imaging modalities and extensive datasets encourage broader applicability and research in medical artificial intelligence.

The GREEN metric's ability to maintain robust performance on OOD data further signifies its versatility and potential as a standard for future developments in automated radiology reporting.

\section{Limitations}
Analyzing each sample takes roughly 3.75 seconds on one A100 GPU. 
However, using batching can accelerate the processing to four samples in 4.22 seconds (equivalent to about 1.06 seconds per sample). Due to its complexity, it is slower compared to ROUGE, at approximately ~0.015 seconds per sample, but faster than GPT-4, at up to 22.0 seconds per sample).

We introduce OOD metrics and suggest a strategy to adjust the GREEN model to different imaging techniques, even in the absence of an initial RRG model for each technique. Nonetheless, fine-tuning GREEN for new imaging modalities might be required in subsequent studies to ensure satisfactory performance.

Although the model operates deterministically to ensure reproducible outputs, the error quantification remains, to some extent, uncontrollable, which introduces a degree of randomness to the counting of errors. This randomness may stem from inherent uncertainties in the task, as evidenced by the disagreement among experts on fine-grained error counts (Section~\ref{sec:interexpert_analysis}). This is a characteristic that has been previously observed and noted in inter-expert agreement analyses~\cite{irvin2019chexpert}. 

\section*{Acknowledgments}
S.O. receives research support by the German Research Foundation ((ID: 517316550). A.C. receives research support from R01 HL167974, R01HL169345, R01 AR077604, R01 EB002524, R01 AR079431, P41 EB027060, AY2AX000045, and contracts 75N92020C00008, 75N92020C00021. C.B. receives research support from Promedica Foundation, Switzerland. \\

This work was also supported in part by the Medical Imaging and Data Resource Center (MIDRC), which is funded by the National Institute of Biomedical Imaging and Bioengineering (NIBIB) of the National Institutes of Health under contract 75N92020D00021 and through The Advanced Research Projects Agency for Health (ARPA-H).

\bibliography{custom}

\appendix
\onecolumn % Switch to one-column layout
\section{Appendix}
\subsection{GPT-4 Prompt Template for Generation of Training Data}
\label{app:gpt4prompttraining}
The following prompt was used in GPT-4 to generate the GREEN model training data. **Reference Report** and **Candidate Report** fields are replaced with their respective actual reports.
\begin{figure}[h]
\begin{AIbox}{GPT-4 Prompt}
{
\fontsize{10}{9} % Set the font size to 12pt with 14pt line spacing
\begin{verbatim}
Objective:
Evaluate the accuracy of a candidate radiology report in comparison to a reference 
radiology report composed by expert radiologists.

Process Overview:
You will be presented with:
1. The criteria for making a judgment.
2. The reference radiology report.
3. The candidate radiology report.
4. The desired format for your assessment.

1. Criteria for Judgment:
For each candidate report, determine:
    - The count of clinically significant errors.
    - The count of clinically insignificant errors.

Errors can fall into one of these categories:
    a) False report of a finding in the candidate.
    b) Missing a finding present in the reference.
    c) Misidentification of a finding's anatomic location/position.
    d) Misassessment of the severity of a finding.
    e) Mentioning a comparison that isn't in the reference.
    f) Omitting a comparison detailing a change from a prior study.

Note: Concentrate on the clinical findings rather than the report's writing style. 
Evaluate only the findings that appear in both reports.

2. Reference Report:
**Reference Report**

3. Candidate Report:
**Candidate Report**

4. Reporting Your Assessment:
Follow this specific format for your output, even if no errors are found:
```
[Explanation]:
<Explanation>

[Clinically Significant Errors]:
(a) <Error Type>: <The number of errors>. <Error 1>; <Error 2>; ...; <Error n>
....
(f) <Error Type>: <The number of errors>. <Error 1>; <Error 2>; ...; <Error n>

[Clinically Insignificant Errors]:
(a) <Error Type>: <The number of errors>. <Error 1>; <Error 2>; ...; <Error n>
....
(f) <Error Type>: <The number of errors>. <Error 1>; <Error 2>; ...; <Error n>

[Matched Findings]:
<The number of matched findings>. <Finding 1>; <Finding 2>; ...; <Finding n>
```
\end{verbatim}
}
\end{AIbox}
\end{figure}

\clearpage

\subsection{GPT-4 Response used as Training Data} \label{app:gpt4radgraphpermutation}

The following depicts a sample GPT-4 response to the prompt in Appendix~\ref{app:gpt4prompttraining} for a pair of reports selected from a 'RadGraph Permutation'. Specifically, the candidate report contains exactly one RadGraph modifier change.\\

\textbf{Reference:} "Examination. Faint infiltrates in the upper middle right field and doubtful retrocardiac suggestive of respiratory infection. Costophrenic sinuses are clear. No other notable findings."

\textbf{Candidate:} "Examination . Subpleural infiltrates in the upper dorsal right field and doubtful retrocardiac suggestive of respiratory infection . Costophrenic sinuses are clear . No other notable findings ."

\begin{figure*}[h]
\begin{AIbox}{GPT-4 Response}
\begin{verbatim}
[Explanation]:
The candidate report misidentifies the anatomic location of the infiltrates. 
The reference report mentions "upper middle right field" while the candidate report 
mentions "upper dorsal right field". 

[Clinically Significant Errors]:
(c) Misidentification of a finding's anatomic location/position: 1. The infiltrates 
are in the upper middle right field, not the upper dorsal right field.

[Clinically Insignificant Errors]:
(a) False report of a finding in the candidate: 0.
(b) Missing a finding present in the reference: 0.
(c) Misidentification of a finding's anatomic location/position: 0.
(d) Misassessment of the severity of a finding: 0.
(e) Mentioning a comparison that isn't in the reference: 0.
(f) Omitting a comparison detailing a change from a prior study: 0.

[Matched Findings]:
3. Doubtful retrocardiac suggestive of respiratory infection; 
Costophrenic sinuses are clear; No other notable findings.
```
\end{verbatim}
\end{AIbox} 
\end{figure*}

\clearpage

\subsection{GPT-4 Prompt Template for Preference Evaluation}
\label{app:gpt4promptpreference}

The following prompt was used in GPT-4 for the evaluation of preferences. **Reference Report**, **Candidate Report 1**, and **Candidate Report 2** fields are replaced with their respective actual reports.

\begin{figure*}[ht]
\begin{AIbox}{GPT-4 Prompt}
\begin{verbatim}
We would like to request your feedback on the radiology reports generated by two AI 
assistants by comparing them to the reference report written by radiologists.

[Reference Report]
**Reference Report**

[Assistant 1]
**Candidate Report 1**

[Assistant 2]
**Candidate Report 2**

[Requirements]
1. The length of the reports is not important.
2. The style of the reports is not important.
3. The clinical accuracy is important especially for positive findings (i.e., diseases).
Therefore, please focus on clinical accuracy instead of the length and style.

Please compare the accuracy of their generated reports. You should tell me whether Assistant 1 
is "better than", "worse than", or "equal to" Assistant 2.

Please first compare the generated reports with the reference report to analyze which one is
more in line with the given requirements.

In the last line, please output a single line containing only a single label selecting from 
"Assistant 1 is better than Assistant 2", "Assistant 1 is worse than Assistant 2", and 
"Assistant 1 is equal to Assistant 2".
\end{verbatim}
\end{AIbox} 
\end{figure*}

\clearpage

\subsection{Algorithm for Modifying Radiology Reports}
\label{app:changereportprompt}
We employed this algorithm to produce prompts for GPT-4 to modify candidate reports for new imaging modalities that lack RRG models.
\begin{algorithm}
\begin{center}
\begin{lstlisting}[language=Python, numbers=none,
                   breaklines=true, linewidth=0.99\textwidth]
def get_prompt(self, report):
        error_types = self.get_error_combination(report)
        # randomly choose if subtle or not, if subtle add "sentence"
        subtle_change = ""
        if random.random() > 0.5:
            subtle_change = "Aim for subtlety, adjusting only one word where feasible. "
        if not error_types == "no errors":
            return f"[Objective]: Create a candidate radiology report that subtly integrates specific errors based on the provided reference report.
            Process Overview: You will be presented with:  
            1. Style of errors.
            2. A reference radiology report to base your candidate report on.
            3. The desired format for your candidate report. Note: Be short in your response! 
            
            Style of errors:
            Introduce errors related to {error_types}. The errors should be woven into the report as if they were genuine observations from a medical image, without any meta-commentary on their accuracy. {subtle_change}
            
            Reference Report: \n{report}\n Desired format for your candidate report:  \n\n [Candidate]: <Candidate Report>"

        return f"[Objective]: Create a candidate radiology report that has the same 
        clinical meaning but is slightly rephrased. 
        Process Overview: You will be presented with:  \n 1.A reference radiology report to base your candidate report on. \n 2. The desired format for your candidate report. Note: Be short in your response! \n\n Reference radiology report: \n{report}\n\n Desired format for your candidate report:  \n\n [Candidate]: <Candidate Report>"
\end{lstlisting}
\end{center}
\end{algorithm}

\clearpage
\subsection{Visualization of the GREEN Summary Clustering Technique}
\label{app:clustering}
Visualization (t-SNE) of the clustering technique used in the GREEN summary. Sentences were clustered for each error subcategory.
\begin{figure}[ht]
    \centering
    \includegraphics{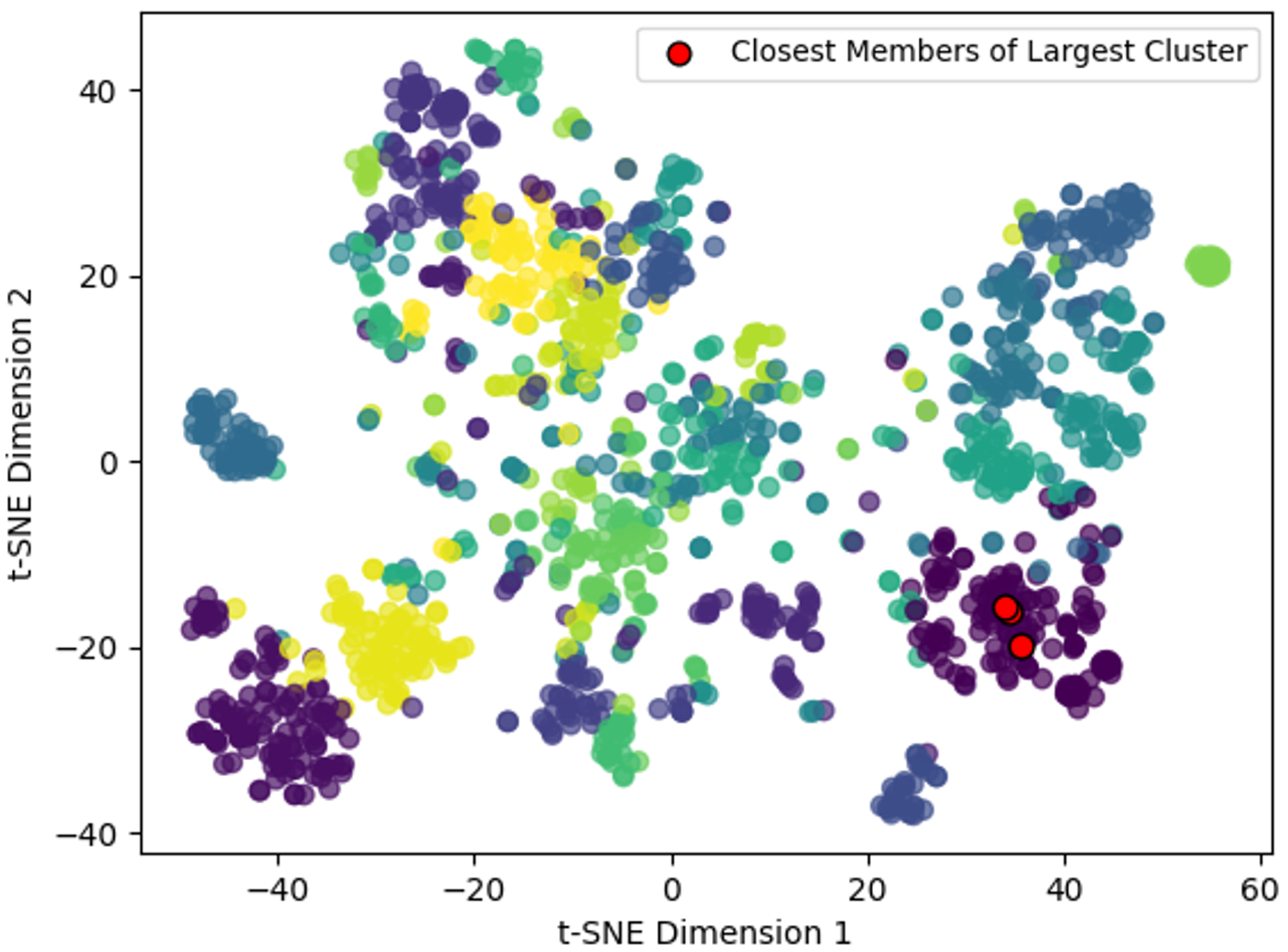}
\end{figure}

\subsection{Test for difference in location of the mean errors}

\begin{table}[h]
\centering
\caption{Paired Wilcoxon test for a significant difference in error counts of experts to the mean expert}
\label{tab:rater_stats}
\begin{tabular}{@{}lcc@{}}
\toprule
Expert    & W Statistic & P-value          \\ \midrule
Expert 0 & 906.0       & $1.83 \times 10^{-18}$ \\
Expert 1 & 1408.0      & $4.77 \times 10^{-14}$ \\
Expert 2 & 2265.0      & $6.22 \times 10^{-9}$  \\
Expert 3 & 2450.0      & $2.01 \times 10^{-8}$  \\
Expert 4 & 3130.0      & $4.76 \times 10^{-5}$  \\
Expert 5 & 1505.5      & $2.11 \times 10^{-13}$ \\ \bottomrule
\end{tabular}
\end{table}

\begin{table}[h]
\centering
\caption{Paired Wilcoxon test for a significant difference in error counts of models to the mean expert}
\label{tab:model_stats}
\begin{tabular}{@{}lcc@{}}
\toprule
Model            & W Statistic & P-value                \\ \midrule
Mistral-v0.1 (7B) & 722.50      & $4.30 \times 10^{-26}$ \\
LaMA-2 (7B)      & 2324.50     & $3.41 \times 10^{-19}$ \\
Phi-2 (2.7B)     & 2294.50     & $4.00 \times 10^{-19}$ \\
RadLLaMA-2 (7B)  & 2345.50     & $2.19 \times 10^{-9}$  \\
RadPhi-2 (2.7B)  & 1728.00     & $4.63 \times 10^{-20}$ \\
GREEN GPT-4      & 2554.00     & $8.93 \times 10^{-9}$  \\ \bottomrule
\end{tabular}
\end{table}

% \begin{table}[H]
% \centering
% \caption{Accuracy of experts to the expert mean}
% \begin{tabular}{lcccccc}
% \toprule
% Model & (a) & (b) & (c) & (d) & (e) & (f) \\
% \midrule
% Expert 0  & 0.40 & 0.49 & 0.66 & 0.60 & 0.68 & 0.7 \\
% Expert 1  & 0.38 & 0.51 & 0.65 & 0.60 & 0.68 & 0.7 \\
% Expert 2  & 0.39 & 0.50 & 0.66 & 0.60 & 0.68 & 0.7 \\
% Expert 3  & 0.39 & 0.49 & 0.65 & 0.61 & 0.68 & 0.7 \\
% Expert 4 & 0.37 & 0.50 & 0.65 & 0.60 & 0.68 & 0.7 \\
% Expert 5  & 0.36 & 0.48 & 0.66 & 0.60 & 0.68 & 0.7 \\\midrule
% GREEN &  0.34	&0.38	&0.60	&0.54	&0.65	&0.68 \\
%  GREEN GPT-4  & 0.32 &  0.40 &  0.65 &  0.59 &  0.68 &  0.70\\  
% \bottomrule
% \end{tabular}
% \end{table}
\subsection{Fine-grained interexpert correlation}

\begin{figure}
    \centering
    \includegraphics[width=0.4\textwidth]{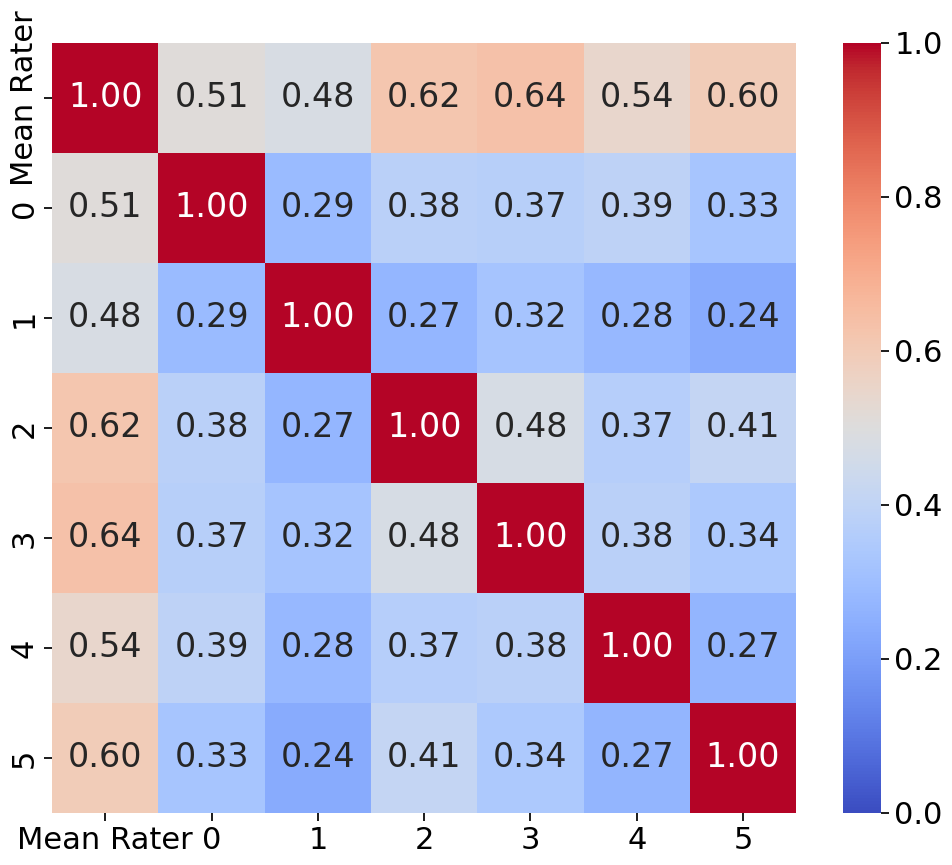}
    \caption{Mean-expert and inter-expert correlation matrix (Kendall's Tau) for fine-grained error counts on the external validation set (RexVal~\cite{yu2023radiology}).}
    \label{fig:inter-expert_corr}
\end{figure}
\end{document}